\pdfoutput=1

\documentclass[11pt]{article}

\usepackage{acl}
\usepackage{times}
\usepackage{latexsym}
\usepackage{algorithm}
\usepackage{algpseudocode}
\usepackage{amsmath}
\usepackage{amssymb}
\usepackage{amsmath}
\usepackage{arydshln}
\usepackage{ascii}
\usepackage{bm}
\usepackage{booktabs}
\usepackage{colortbl}
\usepackage{color}
\usepackage{dsfont}
\usepackage{enumitem}
\usepackage{forest}
\usepackage{graphics}
\usepackage{graphicx}
\usepackage{latexsym}
\usepackage{multirow}
\usepackage{mwe}
\usepackage{makecell}
\usepackage{microtype}
\usepackage{subfig}
\usepackage{subcaption}
\usepackage{times}
\usepackage{tikz}
\usepackage{tabularx}
\usepackage{url}
\usepackage{wrapfig}
\usepackage{tcolorbox}
\usepackage{xcolor}

\definecolor{sred}{RGB}{203, 64, 46}
\definecolor{sblue}{RGB}{44, 73, 135}
\definecolor{sgreen}{RGB}{37, 100, 28}

\definecolor{comments}{RGB}{250, 0, 0}

\usepackage[T1]{fontenc}

\usepackage[utf8]{inputenc}

\usepackage{microtype}

\usepackage{inconsolata}
\usepackage{array} 
\usepackage{lipsum} 
\usepackage{booktabs} 
\usepackage{multirow}
\usepackage{CJKutf8}

\title{LoRA-Flow: Dynamic LoRA Fusion for Large Language Models in Generative Tasks}

\author{
Hanqing Wang$^{*1}$ \
Bowen Ping\thanks{\ \ Equal Contribution.}$^{2}$ \
Shuo Wang$^{\dag3}$ \
Xu Han$^{3}$ \\
\textbf{Yun Chen}\thanks{\ \ Corresponding authors.}$^{1}$ \
\textbf{Zhiyuan Liu}$^{3}$ \
\textbf{Maosong Sun}$^{3}$ \\
$^1$Shanghai University of Finance and Economics \quad $^2$Peking University \\
$^3$Tsinghua University
}

\begin{document}
\maketitle
\begin{abstract}
LoRA employs lightweight modules to customize large language models (LLMs) for each downstream task or domain, where different learned additional modules represent diverse skills. Combining existing LoRAs to address new tasks can enhance the reusability of learned LoRAs, particularly beneficial for tasks with limited annotated data. Most prior works on LoRA combination primarily rely on task-level weights for each involved LoRA, making different examples and tokens share the same LoRA weights. However, in generative tasks, different tokens may necessitate diverse skills to manage. Taking the Chinese math task as an example, understanding the problem description may depend more on the Chinese LoRA, while the calculation part may rely more on the math LoRA. To this end, we propose LoRA-Flow, which utilizes dynamic weights to adjust the impact of different LoRAs. The weights at each step are determined by a fusion gate with extremely few parameters, which can be learned with only 200 training examples. Experiments across six generative tasks demonstrate that our method consistently outperforms baselines with task-level fusion weights. This underscores the necessity of introducing dynamic fusion weights for LoRA combination.\footnote{\ \ Code and models will be publicly available at \url{https://github.com/OpenBMB/LoRAFlow}.}
\end{abstract}

\section{Introduction}

Large language models (LLMs) have demonstrated superior performance over previous smaller models across a wide range of tasks~\cite{openai2023gpt4,anil2023palm,llama,llama2}, thereby extending the applicability of AI systems to numerous real-world scenarios. Because of their substantial model size, training all parameters of LLMs can often be prohibitively expensive. Therefore, several researchers have proposed a set of parameter-efficient fine-tuning (PEFT) approaches~\cite{pmlr-v97-houlsby19a,li-liang-2021-prefix}. Among these, LoRA (low-rank adaptation; \citealt{hu2022lora}) stands out as one of the most popular due to its efficiency and simplicity.

\begin{figure}[t]
    \centering 
    \includegraphics[width=0.9\linewidth]{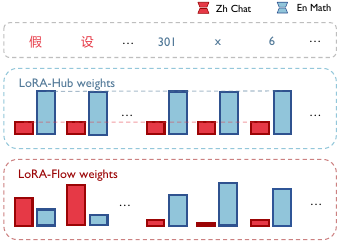}
    \caption{Illustration of the proposed LoRA-Flow method. For the token $y_t$ at the $t$-th step, we use a gate that conditions on the prefix $\mathbf{y}_{<t}$ to determine the fusion weights. The dynamic fusion weights are intended to control the influence of different LoRA modules, to better cope with various types of tokens in generative tasks. Red and blue rectangles represent the weights assigned to the two involved LoRAs.}
    \label{fig:intro}
\end{figure}

The basic idea of LoRA is to learn an additional module for each downstream domain or task. Rather than solely employing a single LoRA to address learned tasks, several recent studies have delved into the potential of combining existing LoRAs to tackle unseen tasks~\cite{zhang2023composing,huang2023lorahub,chronopoulou2023language}. This direction holds the potential to substantially enhance the reusability of learned parameters, facilitating the integration of diverse model capabilities.

Most existing LoRA fusion approaches employ a task-level weight distribution when combining different LoRAs. This implies that all test examples and tokens share the same fusion ratio. However, for some complex generative tasks (e.g., solving mathematical problems or generating code according to provided instructions), the LLM may need to dynamically employ various types of capabilities to address the entire problem effectively. Figure~\ref{fig:intro} illustrates an example, where we have trained a Chinese chat LoRA (i.e., Zh Chat) and an English math LoRA (i.e., En Math), and our objective is to address a Chinese math problem. Intuitively, comprehending the Chinese problem description may rely more on the Chinese chat LoRA, whereas performing the calculation might depend more on the English math LoRA.

In this work, we propose LoRA-Flow, which can dynamically determine the token-level weights of different LoRAs for generative tasks. At each time step, the fusion weights are generated by a gate module that conditions the current prefix. The fusion gate comprises an extremely small number of parameters, accounting for only approximately 0.2\% of those in a LoRA. We find through experiments that the fusion gate can be learned through only 200 training examples. Figure~\ref{fig:intro} gives an example of LoRA-Flow. We also observe significant variations in weights across different model layers, suggesting that the impact of LoRAs differs across layers.
In summary, the contributions of this work can be outlined as follows:
\begin{itemize}
    \item We propose LoRA-Flow, a method that combines existing LoRAs with dynamic fusion weights to effectively control the influence of each LoRA across various generation steps.
    \item We verify the effectiveness of LoRA-Flow on six different generation tasks, and the results show that LoRA-Flow can consistently outperform the baselines that use task-level fusion weights (e.g., LoRA-Hub).
    \item By carefully designed analyses from various aspects, we provide deeper insights into the integration of LoRAs. We consider this journey to be fruitful in constructing a flexible plug-and-play community for LLMs, enabling developers to leverage plugins created by others to build up their own LLM applications.
\end{itemize}

\begin{figure*}[t]
    \centering
    \includegraphics[width=0.9\linewidth]{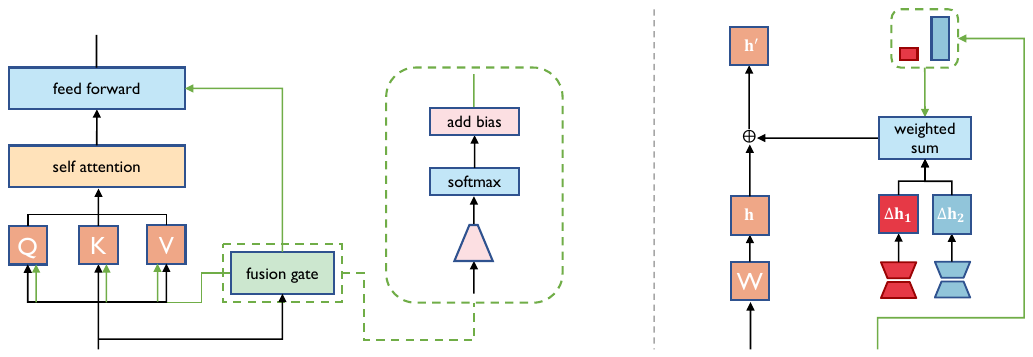}
    \caption{Left: we use layer-wise fusion gates to facilitate dynamic LoRA fusion, which project input hidden states of each layer into fusion weights. Right: for a certain module, the provided fusion weights are used to aggregate the outputs of different LoRAs. Since our goal is to leverage the abilities acquired by existing LoRAs to address new tasks, we only train the fusion gate with a few examples, while keeping both the model and the LoRAs frozen. The number of parameters of the fusion gate is only approximately 0.2\% of those in a LoRA.}
    \label{fig:method}
\end{figure*}

\section{Background}
\label{sec:background}
\paragraph{Large Language Models}
Most recent LLMs utilize a decoder-only architecture, comprising stacked layers of identical structure to form the large-scale model. For a sequence $\mathbf{y}$, an LLM estimates its probability in the following way: 
\begin{equation}
    P(\mathbf{y}|\bm{\theta}_{\mathrm{base}}) = \prod^T_{t=1}P(y_t|\mathbf{y}_{<t};\bm{\theta}_{\mathrm{base}}),
\end{equation}
where $y_t$ denotes the token at the $t$-th step and $\mathbf{y}_{<t}$ is the prefix before $t$. $\bm{\theta}_{\mathrm{base}}$ represents the parameters of the basic LLM.

\paragraph{LoRA}
LoRA~\cite{hu2022lora} is a parameter-efficient fine-tuning method that can achieve comparable performance to full fine-tuning in certain scenarios but at a significantly lower cost. Specifically, for a given matrix $W \in \mathbb{R}^{m \times d}$ within the model, we can learn two low-rank matrices $A \in \mathbb{R}^{r \times d}$ and $B \in \mathbb{R}^{m \times r}$ to approximate the parameter update for $W$:
\begin{equation}
    \Delta W = B A.
\end{equation}

\paragraph{LoRA Fusion} Each trained LoRA possesses unique capabilities, and their combination can integrate various skills of LLMs. Formally, we use $\Delta W_1 = B_1 A_1$ and $\Delta W_2 = B_2 A_2$ to represent two existing LoRAs. \citet{zhang2023composing} propose to merge two LoRAs in the following way:
\begin{equation}
    \mathbf{h}^{\prime} = W\mathbf{x} + \lambda \Delta W_1 \mathbf{x} + (1 - \lambda) \Delta W_2 \mathbf{x},
\end{equation}
where $\lambda$ is a hyper-parameter that needs to be manually tuned.

LoRA-Hub~\cite{huang2023lorahub} further improve the fusion method in the following ways:
\begin{equation}
\begin{split}
    \mathbf{h}^{\prime} &= W\mathbf{x} \\
    &+ (w_1 B_1 + w_2 B_2) (w_1 A_1 + w_2 A_2) \mathbf{x},
\end{split}
\label{eq:lora-hub}
\end{equation}
where the fusion weights $w_1$ and $w_2$ are learned in a few-shot manner. While LoRA-Hub can automatically determine fusion weights for different LoRAs, the weights across different tokens remain the same for a given task. This shared weight scheme may constrain the expressive capacity of the involved LoRAs, particularly in complex generative tasks that entail diverse types of context.

\section{Approach}

To handle generative tasks more flexibly, we propose LoRA-Flow, which employs dynamic fusion weights at each generation step.
Subsequently, we will first introduce the way we calculate the fusion weights in Section~\ref{sec:calc-weight}, and then detail how we integrate the weights into the model in Section~\ref{sec:integrate-weight}. Finally, we describe the training algorithm in Section~\ref{sec:training}.

\subsection{Calculating Fusion Weights}
\label{sec:calc-weight}

At the $t$-th step, we aim to determine the fusion weights using the prefix $\mathbf{y}_{<t}$, which captures the context of the current token. Given that the backbone model has already compressed the context information into hidden vectors, we propose directly utilizing the hidden state at the $t$-th step to avoid redundant computations.

There are three levels of hidden states, each containing different granularities of information:
\begin{itemize}
    \item Step-level hidden states $\mathbf{x}_t$: the input word embedding at the $t$-th step.
    \item Layer-level hidden states $\mathbf{x}_{t}^{l}$: the input representation to the $l$-th layer at the $t$-th step.
    \item Module-level hidden states $\mathbf{x}_{t}^{l, \mathrm{type}}$: the input vector to a specific module (i.e., the query projection in the self-attention network).
\end{itemize}

As found in some previous studies, the hidden states in various layers may lie in different manifolds~\cite{voita-etal-2019-bottom}. Therefore, simply using the step-level hidden state $\mathbf{x}_t$ to compute the fusion weights for the entire model may be not sufficiently effective. On the other side, using module-level hidden states would introduce plenty of new parameters for the fusion gates, as each module requires an independent gate to cope with its input states $\mathbf{x}_{t}^{l, \mathrm{type}}$. We thus use the layer-level hidden state $\mathbf{x}_{t}^{l}$. Empirical studies in Section~\ref{sec:ablation} indicate that layer-wise fusion weights can outperform the other two types of counterparts.

As shown in Figure~\ref{fig:method}, for the $l$-th layer, the fusion gate takes in the input representation $\mathbf{x}_{t}^{l}$ and then projects it into fusion weights:
\begin{equation}
    \mathbf{w}^{l} = \mathrm{softmax}\left(
    W_{\mathrm{gate}}^{l} \mathbf{x}_{t}^{l}
    \right ) + \mathbf{b}^{l},
    \label{eq:gate}
\end{equation}
where $W_{\mathrm{gate}}^{l} \in \mathbb{R}^{k \times d}$ and $\mathbf{b}^{l} \in \mathbb{R}^{k\times1}$. $k$ is the number of LoRAs. Both $W_{\mathrm{gate}}^{l}$ and $\mathbf{b}^{l}$ are learnable parameters. 
Given that $k$ is typically substantially smaller than $d$ and $r$, the number of parameters within the fusion gates is negligible compared to that of the LoRA. For instance, in Llama-2-7b~\cite{llama2}, a LoRA for the entire model contains 117.44M parameters with $r=64$, while the gates for combining two LoRAs only consist of 0.26M parameters in total.

\subsection{Integrating Fusion Weights}
\label{sec:integrate-weight}

As mentioned in Section~\ref{sec:calc-weight}, we use layer-wise fusion weights in LoRA-Flow. Once we get the fusion weights at the $l$-th layer, we feed the weights $\mathbf{w}^l$ to all the modules that contain LoRAs. Different from LoRA-Hub~\cite{huang2023lorahub} that separately composes LoRA A matrices and LoRA B matrices (as shown in Eq~(\ref{eq:lora-hub})), we integrate the outputs of different LoRAs, treating each LoRA as a complete module. The reason for this operation is to combine LoRAs with different middle ranks. Let $\Delta \mathbf{h} = [\Delta \mathbf{h}_1;\cdots;\Delta \mathbf{h}_k] \in \mathbb{R}^{d \times k}$ denote the outputs of all the involved LoRAs, the fusion process can be expressed by
\begin{equation}
    \mathbf{h}^{\prime} = \mathbf{h} + \Delta \mathbf{h} \mathbf{w}^{l},
\end{equation}
where $\mathbf{h}$ is the module’s output in the backbone. Figure~\ref{fig:method} shows an example.

\subsection{Training}
For a backbone model $\bm{\theta}_{\mathrm{base}}$ and a set of existing learned LoRAs $\bm{\theta}_{\mathrm{LoRA}} = \{ \bm{\theta}_{\mathrm{LoRA}}^{1}, \cdots, \bm{\theta}_{\mathrm{LoRA}}^{k} \}$, we train the fusion gate $\bm{\theta}_{\mathrm{fusion}}$ on the new task:
\begin{equation}
    \hat{\bm{\theta}}_{\mathrm{fusion}} = \mathop{\rm argmax}_{\bm{\theta}_{\mathrm{fusion}}} \left \{
    \mathcal{L}(\bm{\theta}_{\mathrm{total}} | \mathcal{D}_{\rm new})
    \right \},
\end{equation}
where $\bm{\theta}_{\mathrm{total}} = \bm{\theta}_{\mathrm{base}} \cup \bm{\theta}_{\mathrm{LoRA}} \cup \bm{\theta}_{\mathrm{fusion}}$ denotes the total parameters. The likelihood is defined as
\begin{equation}
    \mathcal{L}(\bm{\theta}_{\mathrm{total}} | \mathcal{D}_{\rm new}) = \sum_{i=1}^{N} P(\mathbf{y}_i | \bm{\theta}_{\rm total}),
\end{equation}
where $N$ is the number of training examples on the new task. We follow LoRA-Hub~\cite{huang2023lorahub} to learn the fusion modules in a few-shot manner, where $N$ is set to 200 in our experiments. Since $\bm{\theta}_{\rm fusion}$ consists of only a few parameters, these limited training examples are adequate for learning an effective fusion mechanism.

\label{sec:training}

\section{Experiment}
\subsection{Setup}

\paragraph{Base Model}

We use Llama-2~\cite{llama2} as our base LLM to examine the performance of various LoRA fusion approaches, as it is among the most widely used open-source LLMs. Due to computational constraints, we use Llama-2-7b by default.

\paragraph{LoRA Training}

To conduct LoRA fusion experiments, we first learn several LoRAs on the tasks with sufficient supervised data:
\begin{itemize}
    \item Chinese chat (Zh Chat): we use the data released by \citet{okapi} to learn the Chinese chat LoRA, which is expected to possess the ability to understand and generate Chinese text. There are 52K training examples in total.
    \item Russian chat (Ru Chat): the training data of Russian chat LoRA is also from \citet{okapi}, which consists of 52K training examples in Russian.
    \item Spanish chat (Es Chat): the training data is also from \citet{okapi}, containing 52K training examples in Spanish.
    \item English math (En Math): the training data for English math LoRA is constructed by \citet{metamath}, which is comprised of 395K mathematical problems in English.
    \item English code (En Code): we train the English code LoRA with the Magicoder dataset~\cite{wei2023magicoder}, which consists of 186K code generation problems in English.
\end{itemize}

We integrate LoRAs into the query, key, value, and output projections within attention networks and the three linear projections in feedforward networks. By default, the LoRA rank $r$ is set to 64 and the value of $\alpha$ is set to 16. For the En code LoRA, we set $r$=256, as we have observed that the code LoRA with $r$=64 is inadequate for learning the task effectively. We use the cosine warmup schedule and the peak learning rate is 1e-4. For each task, the LoRA is trained by 3 epochs and the warmup ratio is set to 0.04.

\paragraph{LoRA Fusion}

We evaluate LoRA fusion methods in a few-shot manner. Specifically, we conduct the evaluation on six tasks, including Zh math, Ru math, Es math, Zh code, Ru code, and Es code. For each task, we combine the chat LoRA in the target language and the task LoRA in English. Briefly, we use {\em language LoRA} to represent the target-language chat LoRA and {\em task LoRA} to denote the math or code LoRA trained in English.

For each task, we construct 200 training examples for the few-shot training, which is firstly translated by GPT-3.5 based on the English math or code data and then verified by humans. For fusion experiments on math, we also construct 100 problems as the validation set. For code experiments, we employ humans to create 20 problems as the validation set, since the code generation test examples are more time-consuming to annotate.
Both LoRA-Hub and the proposed LoRA-Flow are trained with the few-shot data. We use the validation to search the training hyperparameters. The search space of the peak learning rate is $\{$1e-3, 1e-4$\}$, and the search space of the batch size is $\{$2, 4, 8$\}$. Each fusion module is trained with 5 epochs, using the few-shot training data.

\paragraph{Evaluation}
For fusion experiments on math, we use MGSM~\cite{mgsm} as the test set, which is a widely used multilingual evaluation benchmark for the math abilities of LLMs. For fusion experiments on code generation, we construct a multilingual version of HumanEval~\cite{humaneval}. The original English problem descriptions in HumanEval are translated by GPT-3.5 into other languages and then verified by humans. We report the accuracy and the pass@1 score on MGSM and HumanEval, respectively.

\begin{table*}[ht]
\centering 
\begin{tabular}{l l rrrr rrrr}
\toprule
\multicolumn{2}{c}{\multirow{2}{*}{\bf Method}} & \multicolumn{4}{c}{\bf MGSM (Math)} & \multicolumn{4}{c}{\bf HumanEval (Code)} \\
\cmidrule(lr){3-6} \cmidrule(lr){7-10}
&& \bf Zh & \bf Ru & \bf Es & \bf Avg. & \bf Zh & \bf Ru & \bf Es & \bf Avg. \\
\midrule
\multicolumn{2}{l}{\textproc{Base Model}} & 4.4 & 3.2 & 2.4 & 3.3 & 0.0 & 0.0 & 2.4 & 0.8 \\
\cmidrule(lr){1-2} \cmidrule(lr){3-6} \cmidrule(lr){7-10}
\multirow{2}{*}{\textproc{Single LoRA}} & \textproc{Lang} & 5.2 & 3.6 & 3.6 & 4.1 & 12.2 & 14.0 & 10.4 & 12.2 \\
& \textproc{Task} & 26.8 & 32.8 & 41.2 & 33.6 & 18.3 & 23.2 & 21.9 & 21.1 \\
\cmidrule(lr){1-2} \cmidrule(lr){3-6} \cmidrule(lr){7-10}
\multirow{3}{*}{\textproc{LoRA Fusion}} & \textproc{Average} & 12.8 & 10.4 & 18.4 & 13.9 & 17.1 & 17.7 & 18.3 & 17.7 \\
& \textproc{LoRA-Hub} & 20.8 & 28.4 & 36.8 & 28.7 & 19.5 & 21.3 & 20.1 & 20.3 \\
& \textproc{LoRA-Flow} & \textbf{33.2} & \textbf{37.6} & \textbf{42.0} & \textbf{37.6} & \textbf{20.7} & \textbf{23.8} & \textbf{23.2} & \textbf{22.6} \\
\bottomrule
\end{tabular}
\caption{Evaluation results on MGSM and HumanEval. ``\textproc{Lang}'' denotes the chat LoRA in the target language and ``\textproc{Task}'' represents the math or code LoRA trained in English. LoRA fusion methods combine the language LoRA and the task LoRA to accomplish the new task. The best score is highlighted in \textbf{bold}.
}
\label{7b}
\end{table*}

\subsection{Main Results}

Table~\ref{7b} shows the results of the involved methods on different generative tasks. For comparison, we also show the performances of the base model and single LoRA.
The task LoRAs trained in English already demonstrate a notable degree of cross-lingual transfer capabilities, outperforming the language LoRAs on non-English tasks. For example, in the Zh math task, the task LoRA achieves an accuracy of 26.8, whereas the language LoRA achieves only 5.2. On the code generation tasks, the task LoRA also significantly outperforms the language LoRAs.

When combining the language and task LoRAs, we compare our method with two baselines that use task-level fusion weights.
The "Average" baseline refers to simply averaging the outputs of the two involved LoRAs.
As introduced in Section~\ref{sec:background}, LoRA-Hub learns task-level fusion weights using few-shot training data, similar to the proposed LoRA-Flow. However, LoRA-Flow utilizes dynamic fusion weights, which vary across different time steps and model layers. We use the open-source code released by \citet{huang2023lorahub} to reimplement LoRA-Hub in our experiments. 

From the experiments, we find that ``Average'' and LoRA-Hub perform even worse than the single task LoRA, demonstrating that combining different LoRAs with static weights is not effective enough for complex generative tasks like solving mathematical problems or generating code segments. Specifically, LoRA-Hub only outperforms the single-task LoRA on the Zh code task.

Trained with the same few-shot training data as LoRA-Hub, LoRA-Flow outperforms the baselines across all six examined tasks. 
For instance, LoRA-Flow achieves a performance improvement of 2.3 compared to LoRA-Hub (22.6 vs. 20.3) on code generation tasks.
This confirms the necessity of employing dynamic weights for generative tasks.

\begin{table}[ht]
\centering
\begin{tabular}{lrr}
\toprule 
\bf Method & \bf MGSM & \bf HumanEval \\ 
\midrule
\textproc{Base Model} & 6.8 & 0.0 \\
\cmidrule(lr){1-1} \cmidrule(lr){2-3}
\textproc{Lang LoRA} & 7.6 & 18.9 \\
\textproc{Task LoRA} & 36.8 & 34.7 \\
\cmidrule(lr){1-1} \cmidrule(lr){2-3}
\textproc{Average}   & 16.4 & 30.4 \\
\textproc{LoRA-Hub}  & 40.0 & 34.2 \\
\textproc{LoRA-Flow} & \textbf{41.2} & \textbf{35.4} \\
\bottomrule
\end{tabular}
\caption{Performance of different fusion methods on Llama-2-13b. The results are reported on the Zh math and Zh code test sets.
}
\label{13b}
\end{table}

\subsection{Results on Larger Model}

To confirm the effectiveness of LoRA-Flow with larger LLMs, we also conduct experiments on Llama-2-13b. Given that larger models require higher training costs, we only train three LoRAs for the 13b model, namely Zh Chat, En Math, and En Code. The results of different fusion approaches on the 13b model are shown in Table~\ref{13b}.
On the larger model, all the methods involved achieve better results than those on the 7b model. Compared to LoRA-Hub, LoRA-Flow exhibits superior performances on both the math and the code tasks.
These results also demonstrate that LoRA-Flow is compatible with models of various sizes.

\begin{figure}[h]
    \centering
    \includegraphics[height=0.4784\linewidth]{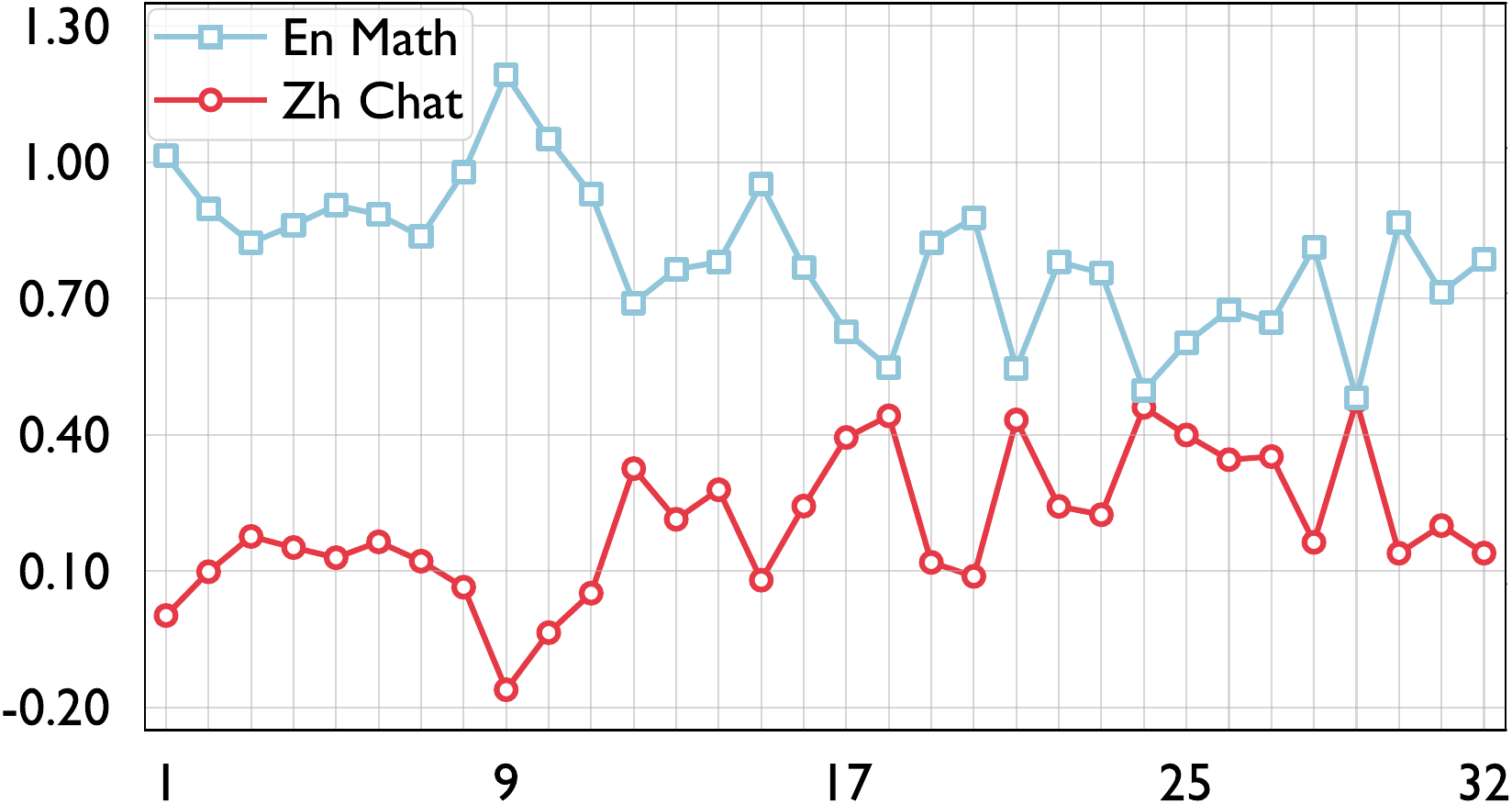}
    \caption{Average fusion weights for the Zh Chat and En Math LoRAs across different layers.
    }
    \label{fig:weight_layer}
\end{figure}

\begin{figure}[h]
    \centering
    \includegraphics[height=0.4784\linewidth]{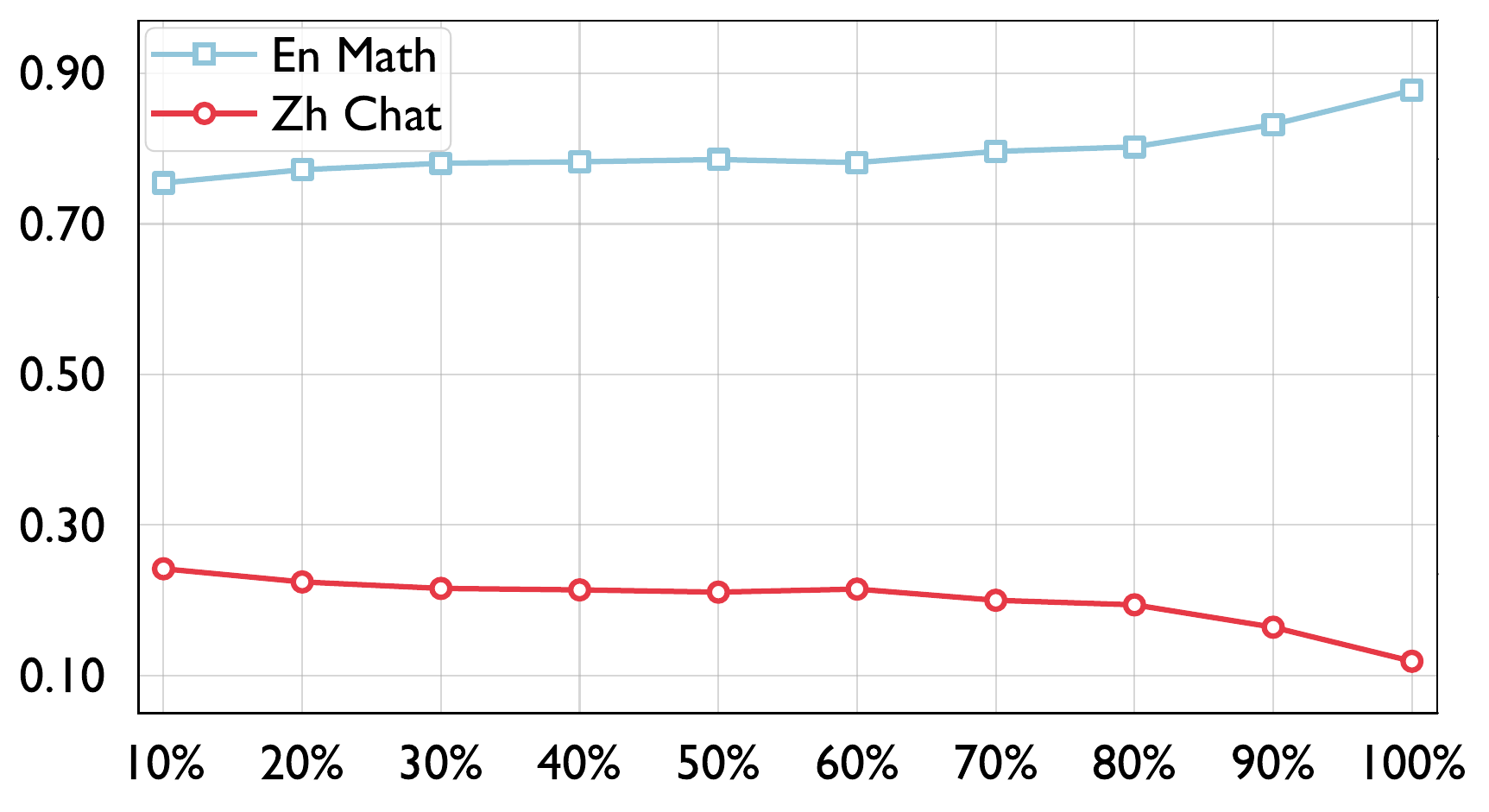}
    \caption{Average fusion weights for the Zh Chat and En Math LoRAs at different time steps.}
    \label{fig:weight_time}
\end{figure}

\section{Analysis}

To gain deeper insights into the behavior of LoRA-Flow, we conduct a comprehensive series of analyses from various perspectives. We first estimate the average fusions at different model layers in Section~\ref{sec:layer_analysis}, and then investigate how the fusion weights change across time steps in Section~\ref{sec:time_analysis}. Finally, we provide a specific case in Section~\ref{sec:case_analysis}.

\begin{figure*}[t]
    \centering
    \includegraphics[width=0.98\linewidth]{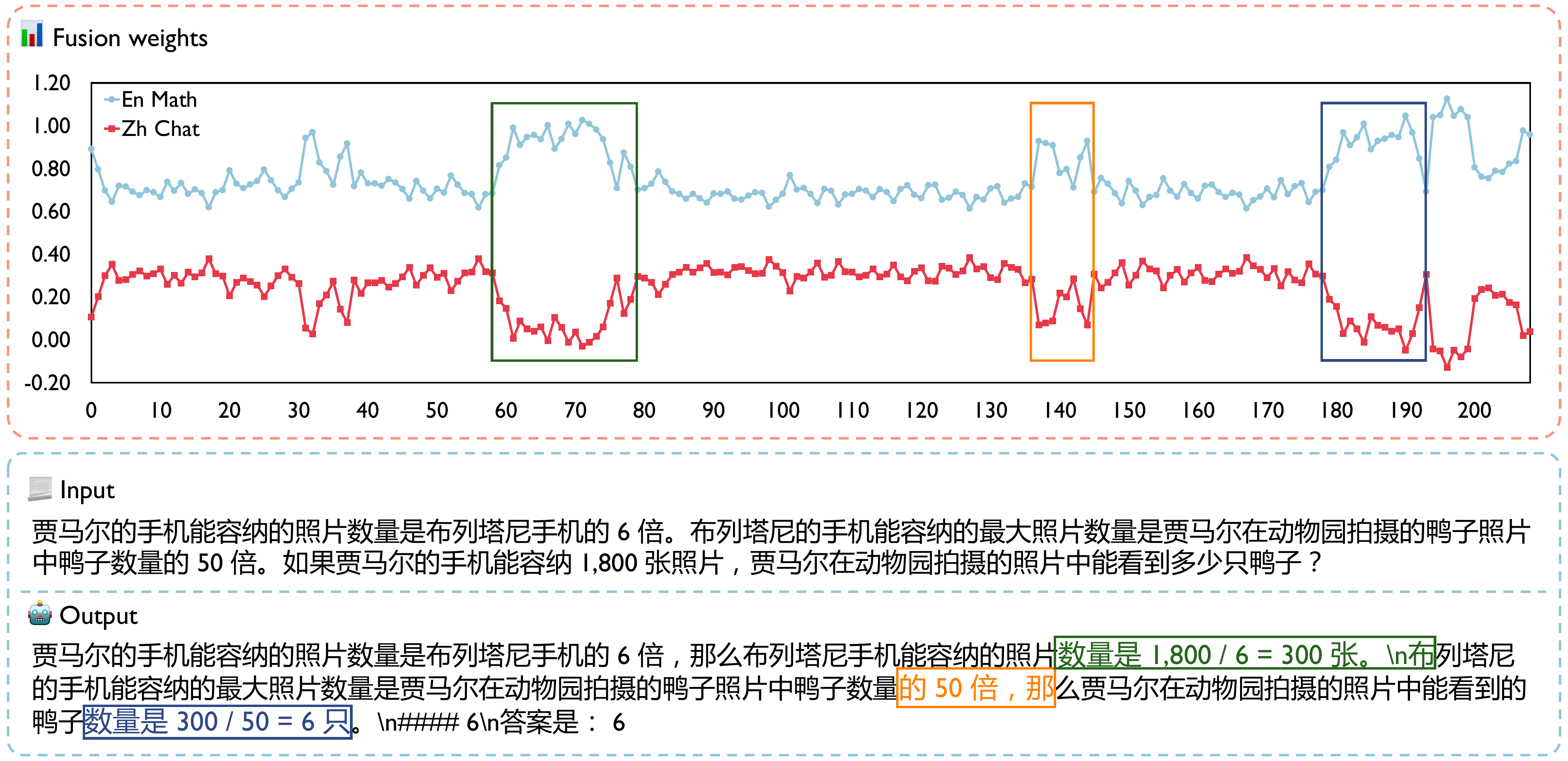}
    \caption{Detailed analysis for the fusion procedure of LoRA-Flow. The upper subgraph illustrates the fusion weights for each token, while the bottom subgraph details the content.
    From the fusion weights, we observe three segments where the fusion weights for the Zh Chat LoRA noticeably decrease while those for the En Math LoRA increase. We highlight the tokens corresponding to these segments using green, yellow, and red colors, respectively. 
    Surprisingly, these three segments mainly contain numbers, which are closely related to mathematical reasoning ability. We also offer English translations of the input and output Chinese text in Figure~\ref{fig:case-study-translated} in the Appendix.
    }
    \label{fig:case-study}
\end{figure*}

\subsection{Fusion Weights across Different Layers}
\label{sec:layer_analysis}
Figure~\ref{fig:weight_layer} shows the average fusion weights at different layers for the two involved LoRAs (i.e., Zh Chat and En Math) on the Chinese math task.
For both the Zh Chat and En Math LoRA, the weights at various layers exhibit notable variations, suggesting distinct fusion schemes across different model layers.
A trend is observed where lower layers assign greater weights to the En Math LoRA, while higher layers allocate more weight to the Zh Chat LoRA. This phenomenon can be explained by the following reason: the bottom layers primarily utilize the math LoRA for reasoning, while the top layers rely more on the language LoRA for text generation in the target language.
In contrast, approaches such as LoRA-Hub, which combines LoRAs with fixed weights, regard the capabilities of each transformer layer as equivalent, overlooking the discrepancies in their abilities. This constrains the potential of LoRA fusion.

\subsection{Fusion Weights at Different Time Steps}
\label{sec:time_analysis}
We also analyze the average fusion weights at different time steps during the decoding process. We split all the tokens generated by the model into 10 bins according to their relative positions. Figure~\ref{fig:weight_time} presents the results, where ``10\%'' denotes the initial 10\% tokens on the left, and ``20\%'' represents the subsequent 10\%-20\% tokens.
The results underscore the necessity for dynamic fusion weights in generative tasks, as they show that different time steps require varying fusion weights, reconfirming our intuition.

\subsection{Detailed Analysis on Fusion Weights}
\label{sec:case_analysis}
To better investigate the fusion procedure of LoRA-Flow, we provide an example in Figure~\ref{fig:case-study}, which consists of the specific tokens generated by the model and the corresponding fusion weights for the two involved LoRAs (i.e., Zh Chat and En Math).
As discussed in Section~\ref{sec:time_analysis}, the fusion weights exhibit significant variation across different tokens. Notably, due to the incorporation of a bias vector in the fusion gate, as depicted in Eq~(\ref{eq:gate}), it is possible for certain tokens to have negative fusion weights. We draw inspiration from the work of \citet{zhang2023composing}, who observed that negative fusion weights can yield specific effects.

On average, the fusion gate assigns higher weights to the En math LoRA. Additionally, we observe certain troughs in the fusion weights for the Zh Chat LoRA. To better understand this phenomenon, we identify the tokens corresponding to these troughs. As shown in Figure~\ref{fig:case-study}, we use the same color to mark the weight throughs and the corresponding text segments. Surprisingly, we observed that when the fusion weight for the Zh chat LoRA decreases, the text segments mainly contain numbers and mathematical calculations (e.g., ``1800 / 6 = 300''), which align more closely with the math capability learned by the En math LoRA.

By examining the case, we observe a strong correlation between the fusion weights with the generated content. When the content involves mathematical reasoning, LoRA-Flow will improve the fusion weight of the En math LoRA. 
The significant fluctuation in weights suggests that relying on fixed weights throughout the decoding process is impractical, underscoring the necessity of dynamic fusion weights.

\section{Discussion}
\subsection{Ablation Study}
\label{sec:ablation}
\begin{table}[ht]
\centering
\begin{tabular}{lcccc}
\toprule
\multirow{2}{*}{\textbf{Method}} & \multicolumn{4}{c}{\textbf{MGSM}}  \\ \cmidrule(lr){2-5}
& \textbf{Zh} & \textbf{Ru} & \textbf{Es}  & \textbf{Avg.}  \\
\midrule
\textproc{Task LoRA}             & 26.8   & 32.8  & 41.2  & 33.6          \\
\textproc{LoRA-Hub} & 20.8 & 28.4 & 36.8 & 28.7 \\
\cmidrule(lr){1-1} \cmidrule(lr){2-5}
\textproc{Step-Level} & 30.0 & 32.4 & \textbf{44.0} & 35.5 \\
\textproc{Layer-Level} & \textbf{33.2} & \textbf{37.6} & 42.0 & \textbf{37.6} \\
\textproc{Module-Level} & 30.4 & 34.0 & 42.4 & 35.6 \\
\bottomrule
\end{tabular}
\caption{Ablation study on various levels of fusion gates.}
\label{ablation}
\end{table}

To further investigate the granularity of the gates in LoRA-Flow, we conduct an ablation study to compare the performance of different types of gates: step-level, layer-level, and module-level gates. As explained in Section~\ref{sec:calc-weight}, the step-level gate estimates the fusion weight for each step, which is shared by the entire model. Layer-level gates represent the default setting in LoRA-Flow, computing the fusion weight at each layer. Module-level gates provide a specific fusion weight for each module (e.g., the query projection module in the last layer), offering greater flexibility than layer-level gates but introducing more trainable parameters.

Table~\ref{ablation} presents the results of varying gate levels. On average, layer-level fusion gates achieve the highest scores, surpassing both step-level and module-level gates. Nevertheless, the other two fusion gate types also outperform LoRA-Hub. Specifically, the step-level fusion gate achieves an average score of 35.5, while that of LoRA-Hub is 28.7. These results suggest that while utilizing shared fusion weights across different model layers, employing dynamic fusion methods at various time steps yields superior results compared to using task-level static weights. Since different model layers may have different capabilities, using layer-level fusion gates can further improve the performance.

\begin{table}[ht]
\centering
\begin{tabular}{lcc}
\toprule
\bf Method & \bf MGSM &\bf HumanEval \\
\midrule
\textproc{FT New LoRA} & 18.8 & 12.2 \\
\cmidrule(lr){1-1} \cmidrule(lr){2-3}
\textproc{FT Lang LoRA} & 16.4 & 15.9 \\
\textproc{FT Task LoRA} & 27.6 & 18.9 \\
\cmidrule(lr){1-1} \cmidrule(lr){2-3}
\textproc{LoRA-Flow} & \textbf{33.2} & \textbf{20.7} \\ 
\bottomrule
\end{tabular}
\caption{Comparison between few-shot fine-tuning and LoRA fusion. The results are reported on the Zh math and Zh code tasks.
}
\label{few-shot}
\end{table}

\subsection{LoRA Fusion vs. Few-shot Fine-Tuning}

In scenarios with limited data, combining existing LoRAs can effectively leverage the knowledge acquired in other tasks. In this section, we compare LoRA-Flow with other few-shot fine-tuning baselines: (1) FT new LoRA: using the same few-shot training examples to learn a new LoRA for the target task; (2) FT lang LoRA: fine-tuning the language LoRA on the target task; and (3) FT task LoRA: fine-tuning the task LoRA on the target task.
The results are shown in Table~\ref{few-shot}.
Our method, which dynamically integrates the language and task LoRAs, surpasses all three few-shot fine-tuning baselines, indicating the effectiveness of utilizing existing LoRAs in few-shot scenarios.

\subsection{Generalization across Different Tasks} 
\label{sec:general}
\begin{figure}[ht]
\centering
\includegraphics[width=\linewidth]{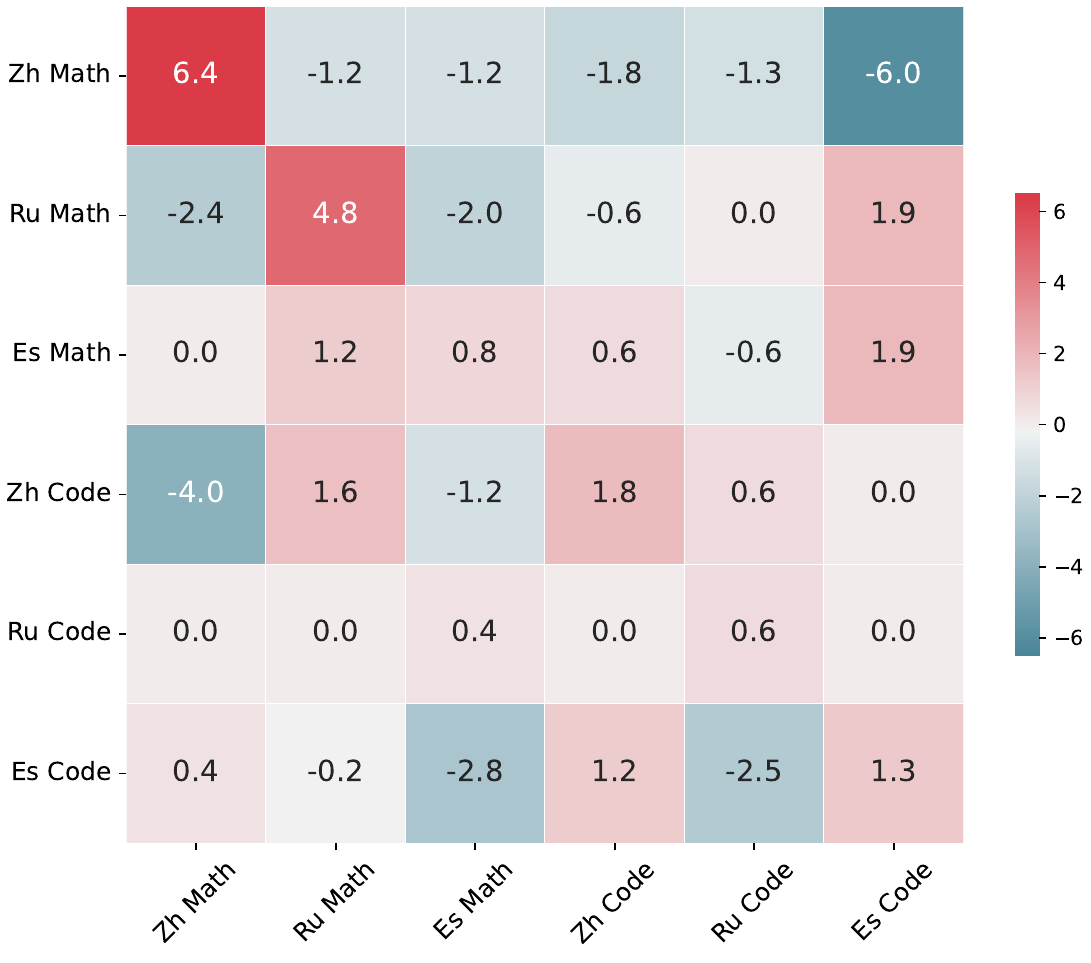}
\caption{Generalizability of the fusion gates. 
Each row represents a training task, and each column represents an evaluation task. The value represents the performance gap between combining the language and task LoRAs and using the task LoRA only. Please refer to Section~\ref{sec:general} for detailed explanations.}
\label{fig:heatmap}
\end{figure}

Similar to LoRA-Hub~\cite{huang2023lorahub}, we learn the fusion module for each task with few-shot training data. In this study, we also explore the generalizability of fusion gates across various tasks. For example, can the gate learned for the Zh code task, which is trained to combine the Zh chat LoRA and the En code LoRA, effectively combine the Ru chat LoRA and the En math LoRA?

Figure~\ref{fig:heatmap} shows the results, depicting the performance difference between merging two LoRAs and employing solely the task LoRA, which is a strong baseline. Specifically, the initial row indicates employing the gate trained with Zh math data for all six generative tasks.
Remarkably, numerous values appear to be positive, suggesting that while zero-shot generalization is not a specific consideration during method design, it still demonstrates some degree of generalization capability. Take the third row as an example, training a gate on the Es math task results in a performance enhancement of 0.8 on the training task. Impressively, this Es math gate also leads to improved performance on three other tasks (i.e., Ru math, Zh code, and Es code). Furthermore, the enhancement observed in the Es code task surpasses that achieved in the Es math task. We leave it to future work to further improve the zero-shot generalization capabilities of LoRA fusion methods.

\section{Related Work}
\paragraph{Module Composition} Exploring the reuse of existing modules for new tasks is an appealing research direction. \citet{pfeiffer2021adapterfusion} combines adapters~\cite{pmlr-v97-houlsby19a} that have been fine-tuned on various downstream tasks through an additional attention layer.
\citet{zhang2023composing} defines some specific operations, such as addition and subtraction, to combine existing LoRAs. The combination weights are tuned on a validation set.
\citet{huang2023lorahub} further improves the LoRA combination by automatically optimizing the fusion weights in a few-shot manner.
\citet{chronopoulou2023language} combines LoRAs trained on the summarization task and multilingual unlabeled data to perform multilingual summarization, where the fusion weight is also a hyperparameter that should be tuned.
Most previous LoRA fusion methods employ task-level static weights, while our proposed LoRA-Flow can dynamically combine different types of LoRA according to the current context.

\paragraph{Mixture-of-LoRA} Some recent studies propose to improve the performance of LoRA with the mixture-of-LoRA (MoLoRA) architecture~\cite{zadouri2023pushing,dou2023loramoe}, which is similar to mixture-of-expert models.
The primary objective of these efforts is to overcome the limitations of LoRA in certain downstream tasks, as the expressive capacity of a single LoRA may be restricted by its intermediate rank.
The major difference between MoLoRA and our work is that MoLoRA mainly aims to train a better plugin for the backbone model, whereas we aim to leverage pre-trained LoRAs for new tasks without training a new LoRA. These two approaches are complementary. Our method can also be extended to incorporate existing MoLoRAs or other advanced types of LoRAs for unseen tasks.

\section{Conclusion}
In this work, we propose LoRA-Flow, a dynamic combination method for LoRA. By assigning dynamic weights to different LoRAs based on the current context, LoRA-Flow can outperform representative baselines with static task-level fusion weights on six complex generative tasks.

\section{Limitation}

In this study, LoRA-Flow outperforms the fixed-weight linear combination method across tasks in various languages. 
However, our computing resources are limited, constraining us to models no larger than 13b parameters.
Nonetheless, the versatility of our method persists, and we leave the exploration of larger models to future work.

\bibliography{custom}

\begin{thebibliography}{19}
\expandafter\ifx\csname natexlab\endcsname\relax\def\natexlab#1{#1}\fi

\bibitem[{Anil et~al.(2023)Anil, Dai, Firat, Johnson, Lepikhin, Passos, Shakeri, Taropa, Bailey, Chen, Chu, Clark, Shafey, Huang, Meier-Hellstern, Mishra, Moreira, Omernick, Robinson, Ruder, Tay, Xiao, Xu, Zhang, Abrego, Ahn, Austin, Barham, Botha, Bradbury, Brahma, Brooks, Catasta, Cheng, Cherry, Choquette-Choo, Chowdhery, Crepy, Dave, Dehghani, Dev, Devlin, Díaz, Du, Dyer, Feinberg, Feng, Fienber, Freitag, Garcia, Gehrmann, Gonzalez, Gur-Ari, Hand, Hashemi, Hou, Howland, Hu, Hui, Hurwitz, Isard, Ittycheriah, Jagielski, Jia, Kenealy, Krikun, Kudugunta, Lan, Lee, Lee, Li, Li, Li, Li, Li, Lim, Lin, Liu, Liu, Maggioni, Mahendru, Maynez, Misra, Moussalem, Nado, Nham, Ni, Nystrom, Parrish, Pellat, Polacek, Polozov, Pope, Qiao, Reif, Richter, Riley, Ros, Roy, Saeta, Samuel, Shelby, Slone, Smilkov, So, Sohn, Tokumine, Valter, Vasudevan, Vodrahalli, Wang, Wang, Wang, Wang, Wieting, Wu, Xu, Xu, Xue, Yin, Yu, Zhang, Zheng, Zheng, Zhou, Zhou, Petrov, and Wu}]{anil2023palm}
Rohan Anil, Andrew~M. Dai, Orhan Firat, Melvin Johnson, Dmitry Lepikhin, Alexandre Passos, Siamak Shakeri, Emanuel Taropa, Paige Bailey, Zhifeng Chen, Eric Chu, Jonathan~H. Clark, Laurent~El Shafey, Yanping Huang, Kathy Meier-Hellstern, Gaurav Mishra, Erica Moreira, Mark Omernick, Kevin Robinson, Sebastian Ruder, Yi~Tay, Kefan Xiao, Yuanzhong Xu, Yujing Zhang, Gustavo~Hernandez Abrego, Junwhan Ahn, Jacob Austin, Paul Barham, Jan Botha, James Bradbury, Siddhartha Brahma, Kevin Brooks, Michele Catasta, Yong Cheng, Colin Cherry, Christopher~A. Choquette-Choo, Aakanksha Chowdhery, Clément Crepy, Shachi Dave, Mostafa Dehghani, Sunipa Dev, Jacob Devlin, Mark Díaz, Nan Du, Ethan Dyer, Vlad Feinberg, Fangxiaoyu Feng, Vlad Fienber, Markus Freitag, Xavier Garcia, Sebastian Gehrmann, Lucas Gonzalez, Guy Gur-Ari, Steven Hand, Hadi Hashemi, Le~Hou, Joshua Howland, Andrea Hu, Jeffrey Hui, Jeremy Hurwitz, Michael Isard, Abe Ittycheriah, Matthew Jagielski, Wenhao Jia, Kathleen Kenealy, Maxim Krikun, Sneha Kudugunta, Chang Lan, Katherine Lee, Benjamin Lee, Eric Li, Music Li, Wei Li, YaGuang Li, Jian Li, Hyeontaek Lim, Hanzhao Lin, Zhongtao Liu, Frederick Liu, Marcello Maggioni, Aroma Mahendru, Joshua Maynez, Vedant Misra, Maysam Moussalem, Zachary Nado, John Nham, Eric Ni, Andrew Nystrom, Alicia Parrish, Marie Pellat, Martin Polacek, Alex Polozov, Reiner Pope, Siyuan Qiao, Emily Reif, Bryan Richter, Parker Riley, Alex~Castro Ros, Aurko Roy, Brennan Saeta, Rajkumar Samuel, Renee Shelby, Ambrose Slone, Daniel Smilkov, David~R. So, Daniel Sohn, Simon Tokumine, Dasha Valter, Vijay Vasudevan, Kiran Vodrahalli, Xuezhi Wang, Pidong Wang, Zirui Wang, Tao Wang, John Wieting, Yuhuai Wu, Kelvin Xu, Yunhan Xu, Linting Xue, Pengcheng Yin, Jiahui Yu, Qiao Zhang, Steven Zheng, Ce~Zheng, Weikang Zhou, Denny Zhou, Slav Petrov, and Yonghui Wu. 2023.
\newblock \href {http://arxiv.org/abs/2305.10403} {Palm 2 technical report}.

\bibitem[{Chen et~al.(2021)Chen, Tworek, Jun, Yuan, de~Oliveira~Pinto, Kaplan, Edwards, Burda, Joseph, Brockman, Ray, Puri, Krueger, Petrov, Khlaaf, Sastry, Mishkin, Chan, Gray, Ryder, Pavlov, Power, Kaiser, Bavarian, Winter, Tillet, Such, Cummings, Plappert, Chantzis, Barnes, Herbert-Voss, Guss, Nichol, Paino, Tezak, Tang, Babuschkin, Balaji, Jain, Saunders, Hesse, Carr, Leike, Achiam, Misra, Morikawa, Radford, Knight, Brundage, Murati, Mayer, Welinder, McGrew, Amodei, McCandlish, Sutskever, and Zaremba}]{humaneval}
Mark Chen, Jerry Tworek, Heewoo Jun, Qiming Yuan, Henrique~Ponde de~Oliveira~Pinto, Jared Kaplan, Harri Edwards, Yuri Burda, Nicholas Joseph, Greg Brockman, Alex Ray, Raul Puri, Gretchen Krueger, Michael Petrov, Heidy Khlaaf, Girish Sastry, Pamela Mishkin, Brooke Chan, Scott Gray, Nick Ryder, Mikhail Pavlov, Alethea Power, Lukasz Kaiser, Mohammad Bavarian, Clemens Winter, Philippe Tillet, Felipe~Petroski Such, Dave Cummings, Matthias Plappert, Fotios Chantzis, Elizabeth Barnes, Ariel Herbert-Voss, William~Hebgen Guss, Alex Nichol, Alex Paino, Nikolas Tezak, Jie Tang, Igor Babuschkin, Suchir Balaji, Shantanu Jain, William Saunders, Christopher Hesse, Andrew~N. Carr, Jan Leike, Josh Achiam, Vedant Misra, Evan Morikawa, Alec Radford, Matthew Knight, Miles Brundage, Mira Murati, Katie Mayer, Peter Welinder, Bob McGrew, Dario Amodei, Sam McCandlish, Ilya Sutskever, and Wojciech Zaremba. 2021.
\newblock \href {http://arxiv.org/abs/2107.03374} {Evaluating large language models trained on code}.

\bibitem[{Chronopoulou et~al.(2023)Chronopoulou, Pfeiffer, Maynez, Wang, Ruder, and Agrawal}]{chronopoulou2023language}
Alexandra Chronopoulou, Jonas Pfeiffer, Joshua Maynez, Xinyi Wang, Sebastian Ruder, and Priyanka Agrawal. 2023.
\newblock \href {http://arxiv.org/abs/2311.09344} {Language and task arithmetic with parameter-efficient layers for zero-shot summarization}.

\bibitem[{Dou et~al.(2023)Dou, Zhou, Liu, Gao, Zhao, Shen, Zhou, Xi, Wang, Fan, Pu, Zhu, Zheng, Gui, Zhang, and Huang}]{dou2023loramoe}
Shihan Dou, Enyu Zhou, Yan Liu, Songyang Gao, Jun Zhao, Wei Shen, Yuhao Zhou, Zhiheng Xi, Xiao Wang, Xiaoran Fan, Shiliang Pu, Jiang Zhu, Rui Zheng, Tao Gui, Qi~Zhang, and Xuanjing Huang. 2023.
\newblock \href {http://arxiv.org/abs/2312.09979} {Loramoe: Revolutionizing mixture of experts for maintaining world knowledge in language model alignment}.

\bibitem[{Houlsby et~al.(2019)Houlsby, Giurgiu, Jastrzebski, Morrone, De~Laroussilhe, Gesmundo, Attariyan, and Gelly}]{pmlr-v97-houlsby19a}
Neil Houlsby, Andrei Giurgiu, Stanislaw Jastrzebski, Bruna Morrone, Quentin De~Laroussilhe, Andrea Gesmundo, Mona Attariyan, and Sylvain Gelly. 2019.
\newblock \href {https://proceedings.mlr.press/v97/houlsby19a.html} {Parameter-efficient transfer learning for {NLP}}.
\newblock In \emph{Proceedings of the 36th International Conference on Machine Learning}.

\bibitem[{Hu et~al.(2022)Hu, yelong shen, Wallis, Allen-Zhu, Li, Wang, Wang, and Chen}]{hu2022lora}
Edward~J Hu, yelong shen, Phillip Wallis, Zeyuan Allen-Zhu, Yuanzhi Li, Shean Wang, Lu~Wang, and Weizhu Chen. 2022.
\newblock \href {https://openreview.net/forum?id=nZeVKeeFYf9} {Lo{RA}: Low-rank adaptation of large language models}.
\newblock In \emph{International Conference on Learning Representations}.

\bibitem[{Huang et~al.(2024)Huang, Liu, Lin, Pang, Du, and Lin}]{huang2023lorahub}
Chengsong Huang, Qian Liu, Bill~Yuchen Lin, Tianyu Pang, Chao Du, and Min Lin. 2024.
\newblock \href {http://arxiv.org/abs/2307.13269} {Lorahub: Efficient cross-task generalization via dynamic lora composition}.

\bibitem[{Lai et~al.(2023)Lai, Nguyen, Ngo, Nguyen, Dernoncourt, Rossi, and Nguyen}]{okapi}
Viet Lai, Chien Nguyen, Nghia Ngo, Thuat Nguyen, Franck Dernoncourt, Ryan Rossi, and Thien Nguyen. 2023.
\newblock \href {https://aclanthology.org/2023.emnlp-demo.28} {Okapi: Instruction-tuned large language models in multiple languages with reinforcement learning from human feedback}.
\newblock In \emph{Proceedings of the 2023 Conference on Empirical Methods in Natural Language Processing: System Demonstrations}.

\bibitem[{Li and Liang(2021)}]{li-liang-2021-prefix}
Xiang~Lisa Li and Percy Liang. 2021.
\newblock \href {https://aclanthology.org/2021.acl-long.353} {Prefix-tuning: Optimizing continuous prompts for generation}.
\newblock In \emph{Proceedings of the 59th Annual Meeting of the Association for Computational Linguistics and the 11th International Joint Conference on Natural Language Processing (Volume 1: Long Papers)}.

\bibitem[{OpenAI(2023)}]{openai2023gpt4}
OpenAI. 2023.
\newblock \href {http://arxiv.org/abs/2303.08774} {Gpt-4 technical report}.

\bibitem[{Pfeiffer et~al.(2021)Pfeiffer, Kamath, R{\"u}ckl{\'e}, Cho, and Gurevych}]{pfeiffer2021adapterfusion}
Jonas Pfeiffer, Aishwarya Kamath, Andreas R{\"u}ckl{\'e}, Kyunghyun Cho, and Iryna Gurevych. 2021.
\newblock Adapterfusion: Non-destructive task composition for transfer learning.
\newblock In \emph{Proceedings of the 16th Conference of the European Chapter of the Association for Computational Linguistics: Main Volume}.

\bibitem[{Shi et~al.(2023)Shi, Suzgun, Freitag, Wang, Srivats, Vosoughi, Chung, Tay, Ruder, Zhou, Das, and Wei}]{mgsm}
Freda Shi, Mirac Suzgun, Markus Freitag, Xuezhi Wang, Suraj Srivats, Soroush Vosoughi, Hyung~Won Chung, Yi~Tay, Sebastian Ruder, Denny Zhou, Dipanjan Das, and Jason Wei. 2023.
\newblock \href {https://openreview.net/forum?id=fR3wGCk-IXp} {Language models are multilingual chain-of-thought reasoners}.
\newblock In \emph{The Eleventh International Conference on Learning Representations}.

\bibitem[{Touvron et~al.(2023{\natexlab{a}})Touvron, Lavril, Izacard, Martinet, Lachaux, Lacroix, Rozière, Goyal, Hambro, Azhar, Rodriguez, Joulin, Grave, and Lample}]{llama}
Hugo Touvron, Thibaut Lavril, Gautier Izacard, Xavier Martinet, Marie-Anne Lachaux, Timothée Lacroix, Baptiste Rozière, Naman Goyal, Eric Hambro, Faisal Azhar, Aurelien Rodriguez, Armand Joulin, Edouard Grave, and Guillaume Lample. 2023{\natexlab{a}}.
\newblock \href {http://arxiv.org/abs/2302.13971} {Llama: Open and efficient foundation language models}.

\bibitem[{Touvron et~al.(2023{\natexlab{b}})Touvron, Martin, Stone, Albert, Almahairi, Babaei, Bashlykov, Batra, Bhargava, Bhosale, Bikel, Blecher, Ferrer, Chen, Cucurull, Esiobu, Fernandes, Fu, Fu, Fuller, Gao, Goswami, Goyal, Hartshorn, Hosseini, Hou, Inan, Kardas, Kerkez, Khabsa, Kloumann, Korenev, Koura, Lachaux, Lavril, Lee, Liskovich, Lu, Mao, Martinet, Mihaylov, Mishra, Molybog, Nie, Poulton, Reizenstein, Rungta, Saladi, Schelten, Silva, Smith, Subramanian, Tan, Tang, Taylor, Williams, Kuan, Xu, Yan, Zarov, Zhang, Fan, Kambadur, Narang, Rodriguez, Stojnic, Edunov, and Scialom}]{llama2}
Hugo Touvron, Louis Martin, Kevin Stone, Peter Albert, Amjad Almahairi, Yasmine Babaei, Nikolay Bashlykov, Soumya Batra, Prajjwal Bhargava, Shruti Bhosale, Dan Bikel, Lukas Blecher, Cristian~Canton Ferrer, Moya Chen, Guillem Cucurull, David Esiobu, Jude Fernandes, Jeremy Fu, Wenyin Fu, Brian Fuller, Cynthia Gao, Vedanuj Goswami, Naman Goyal, Anthony Hartshorn, Saghar Hosseini, Rui Hou, Hakan Inan, Marcin Kardas, Viktor Kerkez, Madian Khabsa, Isabel Kloumann, Artem Korenev, Punit~Singh Koura, Marie-Anne Lachaux, Thibaut Lavril, Jenya Lee, Diana Liskovich, Yinghai Lu, Yuning Mao, Xavier Martinet, Todor Mihaylov, Pushkar Mishra, Igor Molybog, Yixin Nie, Andrew Poulton, Jeremy Reizenstein, Rashi Rungta, Kalyan Saladi, Alan Schelten, Ruan Silva, Eric~Michael Smith, Ranjan Subramanian, Xiaoqing~Ellen Tan, Binh Tang, Ross Taylor, Adina Williams, Jian~Xiang Kuan, Puxin Xu, Zheng Yan, Iliyan Zarov, Yuchen Zhang, Angela Fan, Melanie Kambadur, Sharan Narang, Aurelien Rodriguez, Robert Stojnic, Sergey Edunov, and Thomas Scialom. 2023{\natexlab{b}}.
\newblock \href {http://arxiv.org/abs/2307.09288} {Llama 2: Open foundation and fine-tuned chat models}.

\bibitem[{Voita et~al.(2019)Voita, Sennrich, and Titov}]{voita-etal-2019-bottom}
Elena Voita, Rico Sennrich, and Ivan Titov. 2019.
\newblock \href {https://aclanthology.org/D19-1448} {The bottom-up evolution of representations in the transformer: A study with machine translation and language modeling objectives}.
\newblock In \emph{Proceedings of the 2019 Conference on Empirical Methods in Natural Language Processing and the 9th International Joint Conference on Natural Language Processing (EMNLP-IJCNLP)}.

\bibitem[{Wei et~al.(2023)Wei, Wang, Liu, Ding, and Zhang}]{wei2023magicoder}
Yuxiang Wei, Zhe Wang, Jiawei Liu, Yifeng Ding, and Lingming Zhang. 2023.
\newblock \href {http://arxiv.org/abs/2312.02120} {Magicoder: Source code is all you need}.

\bibitem[{Yu et~al.(2023)Yu, Jiang, Shi, Yu, Liu, Zhang, Kwok, Li, Weller, and Liu}]{metamath}
Longhui Yu, Weisen Jiang, Han Shi, Jincheng Yu, Zhengying Liu, Yu~Zhang, James~T. Kwok, Zhenguo Li, Adrian Weller, and Weiyang Liu. 2023.
\newblock \href {http://arxiv.org/abs/2309.12284} {Metamath: Bootstrap your own mathematical questions for large language models}.

\bibitem[{Zadouri et~al.(2023)Zadouri, Üstün, Ahmadian, Ermiş, Locatelli, and Hooker}]{zadouri2023pushing}
Ted Zadouri, Ahmet Üstün, Arash Ahmadian, Beyza Ermiş, Acyr Locatelli, and Sara Hooker. 2023.
\newblock \href {http://arxiv.org/abs/2309.05444} {Pushing mixture of experts to the limit: Extremely parameter efficient moe for instruction tuning}.

\bibitem[{Zhang et~al.(2023)Zhang, Chen, Liu, and He}]{zhang2023composing}
Jinghan Zhang, Shiqi Chen, Junteng Liu, and Junxian He. 2023.
\newblock Composing parameter-efficient modules with arithmetic operations.
\newblock In \emph{Advances in Neural Information Processing Systems}.

\end{thebibliography}

\appendix

\begin{figure*}[t]
    \centering
    \includegraphics[width=0.98\linewidth]{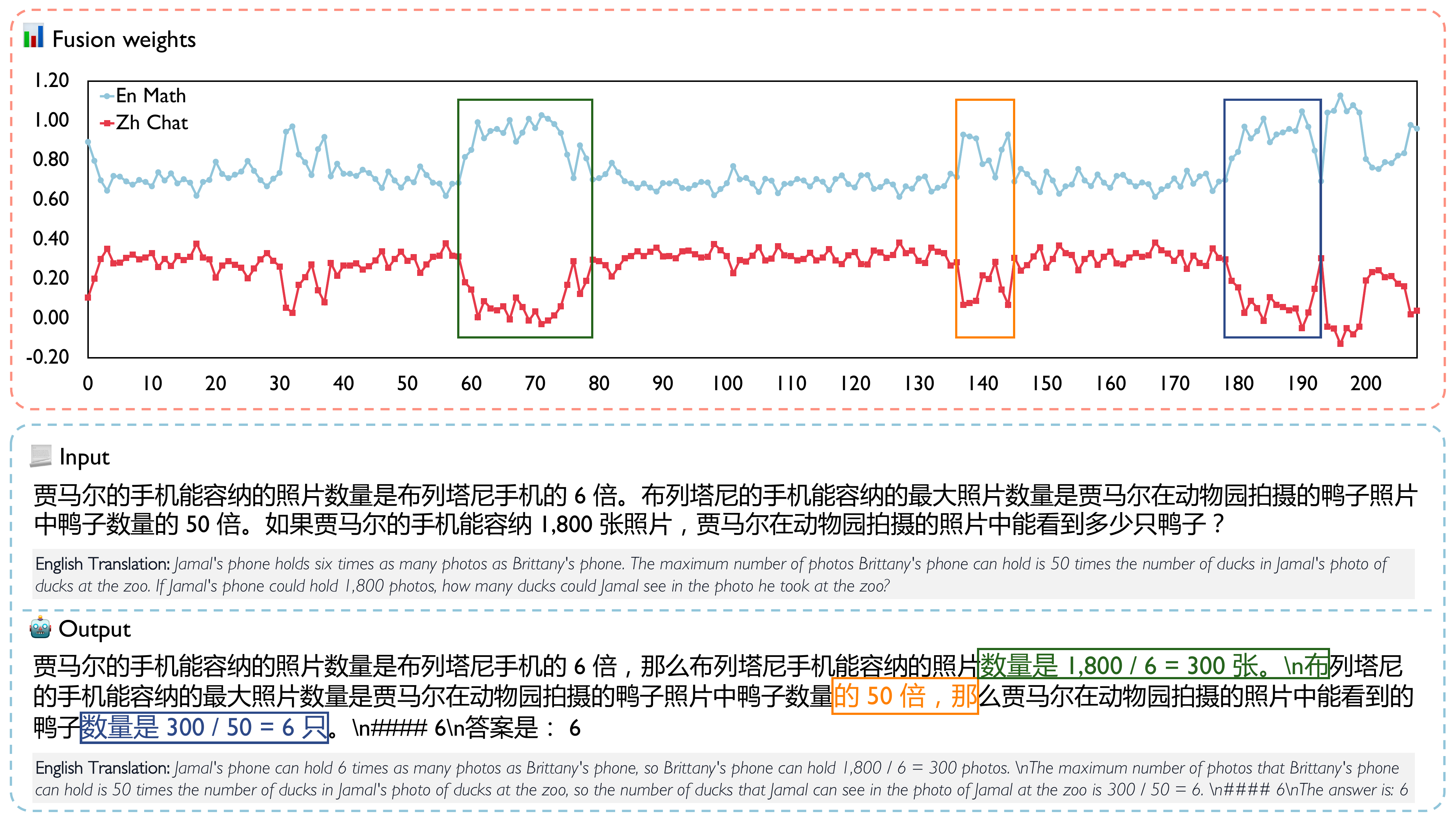}
    \caption{Case study of LoRA-Flow with English translations.}
    \label{fig:case-study-translated}
\end{figure*}

\section{Translated Case}
\label{sec:appendix-case-study}

To better understand the case study presented in Figure~\ref{fig:case-study}, we translate the Chinese question and answer into English, the results are shown in Figure~\ref{fig:case-study-translated}.

\end{document}